\documentclass[10pt,twocolumn,letterpaper]{article}

\usepackage[pagenumbers]{cvpr} 

\usepackage[dvipsnames]{xcolor}

\definecolor{orange}{rgb}{1.0, 0.5, 0.0}
\definecolor{blue-green}{rgb}{0, 0.6, 0.6}
\definecolor{pink}{rgb}{1, 0.2, 0.6}
\definecolor{blue}{rgb}{0.2, 0.6, 1.0}

\newcommand{\mname}[1]{MoRo}
\newcommand{\myparagraph}[1]{\noindent\textbf{#1}}

\usepackage[utf8]{inputenc} 
\usepackage[T1]{fontenc}    
\usepackage{url}            
\usepackage{booktabs}       
\usepackage{amsfonts}       
\usepackage{nicefrac}       
\usepackage{microtype}      
\usepackage{xcolor}         
\usepackage{amsmath}
\usepackage{graphicx}
\usepackage{epsfig}
\usepackage{caption}
\usepackage{subcaption}
\usepackage{multirow}
\usepackage{cuted}
\usepackage{float}

\DeclareMathOperator*{\argmin}{arg\,min}

\definecolor{cvprblue}{rgb}{0.21,0.49,0.74}
\usepackage[pagebackref,breaklinks,colorlinks,citecolor=cvprblue]{hyperref}

\def\paperID{89} 
\def\confName{3DV\xspace}
\def\confYear{2026\xspace}

\title{Masked Modeling for Human Motion Recovery Under Occlusions}

\author{
Zhiyin Qian$^{1*}$ \quad
Siwei Zhang$^{2\dag}$ \quad
Bharat Lal Bhatnagar$^{2}$ \quad
Federica Bogo$^{2}$ \quad
Siyu Tang$^{1}$ \\
{\small $^{1}$ETH Z\"urich} \quad
{\small $^{2}$Meta Reality Labs}\\
} 

\begin{document}

\let\oldtwocolumn\twocolumn
\renewcommand\twocolumn[1][]{%
    \oldtwocolumn[{#1}{
    \begin{center}
       \includegraphics[width=\linewidth]{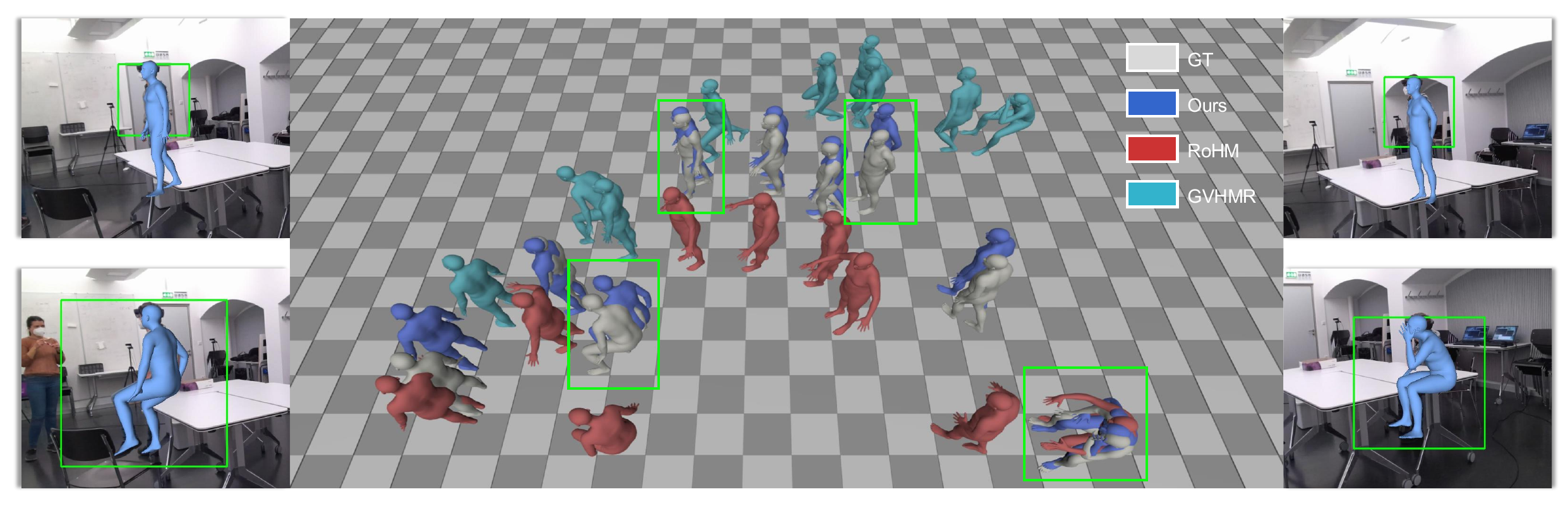}
       \captionof{figure}{\textbf{Overview.} Given a monocular video captured from a static camera, \mname{} robustly reconstructs accurate and physically plausible human motion (middle), even under challenging occlusion scenarios where detections cover only portions of the subject (on both sides). Leveraging masked modeling, our method iteratively synthesizes motion in a globally consistent coordinate by integrating both visual cues and motion priors, facilitated by our cross-modality training strategy. Compared to baselines such as GVHMR~\cite{shen2024world} and RoHM~\cite{zhang2024rohm}, \mname{} consistently achieves better performance in terms of both per-frame accuracy and global motion realism, particularly in the presence of occlusions.
       }
       \label{fig:teaser}
    \end{center}
    }]
}
\maketitle
\def\thefootnote{*}\footnotetext{
All data access, experiments, and model training were conducted at ETH Zürich.}
\def\thefootnote{\dag}\footnotetext{
This work was completed while SZ was a postdoctoral researcher at ETH Zürich.}


\begin{abstract}

    Human motion reconstruction from monocular videos is a fundamental challenge in computer vision, with broad applications in AR/VR, robotics, and digital content creation, but remains challenging under frequent occlusions in real-world settings. 
    Existing regression-based methods are efficient but fragile to missing observations, while optimization- and diffusion-based approaches improve robustness at the cost of slow inference speed and heavy preprocessing steps. 
    To address these limitations, we leverage recent advances in generative masked modeling and present \textbf{MoRo}: \textbf{M}asked m\textbf{O}deling for human motion \textbf{R}ecovery under \textbf{O}cclusions. \mname{} is an occlusion-robust, end-to-end generative framework that formulates motion reconstruction as a video-conditioned task, and efficiently recover human motion in a consistent global coordinate system from RGB videos.
    By masked modeling, MoRo naturally handles occlusions while enabling efficient, end-to-end inference. 
    To overcome the scarcity of paired video–motion data, we design a cross-modality learning scheme that learns multi-modal priors from a set of heterogeneous datasets: (i) a trajectory-aware motion prior trained on MoCap datasets, (ii) an image-conditioned pose prior trained on image-pose datasets, capturing diverse frame-level poses, and (iii) a video-conditioned masked transformer that fuses motion and pose priors, finetuned on video–motion datasets to integrate visual cues with motion dynamics for robust inference.
    Extensive experiments on EgoBody and RICH demonstrate that MoRo substantially outperforms state-of-the-art methods in accuracy and motion realism under occlusions, while performing on-par in non-occluded scenarios. MoRo achieves real-time inference at 70 FPS on a single H200 GPU. 
    Project page: \url{https://mikeqzy.github.io/MoRo}.

\end{abstract}
\section{Introduction}
\label{sec:intro}

Reconstructing 3D human pose and motion from monocular RGB inputs is a long-standing problem in computer vision~\cite{choi2020beyond,cho2022cross,Choi_2020_ECCV_Pose2Mesh,kanazawa2018end,Kocabas_PARE_2021,Kocabas_SPEC_2021,li2021hybrik,li2022cliff,Moon_2020_ECCV_I2L-MeshNet,xu2019denserac,zhang2021body,zhou2021monocular, bogo2016smpl,fang2021reconstructing,SMPL-X:2019, kocabas2020vibe,nam2023cyclic,choi2021beyond, zhang20233d, li2023jotr, khirodkar2022occluded, liu2022explicit, zhang2023probabilistic, Rockwell2020}, with broad applications in augmented and virtual reality, assistive robotics, and healthcare.
However, the limited field of view of monocular cameras often leads to body occlusions when capturing people moving in real-world environments, making motion reconstruction challenging. 
%
Despite the recent rapid progress in this area driven by advances in deep neural network architectures~\cite{vaswani2017attention}, 
existing methods still struggle with such occlusions. 

Most regression-based approaches~\cite{choi2020beyond,kanazawa2019learning,kocabas2020vibe,nam2023cyclic} offer end-to-end fast inference, but perform poorly under heavy occlusions.

Recent methods tackle dynamic camera scenarios~\cite{yuan2022glamr, ye2023slahmr, kocabas2024pace, wham:cvpr:2024, wang2024tram, shen2024world} by jointly estimating human and camera motion in global space, but without explicitly addressing occlusion scenarios. 
Generative modeling is well-suited to tackle the motion ambiguities caused by occlusions.
Optimization-based methods such as HuMoR~\cite{rempe2021humor} and PhaseMP~\cite{phasemp} incorporate VAE-based motion priors within optimization loops~\cite{phasemp, rempe2021humor}, yielding more robust performance than regressors but at the cost of slow inference, sensitivity to initialization, and susceptibility to local minima.
RoHM~\cite{zhang2024rohm} surpasses optimization-based methods in speed and robustness by casting the task as conditional diffusion, but does not provide real-time performance, still fails under severe occlusions, and relies on pose initialization and precomputed body visibility.
Moreover, most of these methods depend solely on precomputed 2D/3D joints and discard the rich visual context available in videos.
These limitations underscore the need for occlusion-robust, end-to-end models capable of real-time inference.

%
To fill this gap we propose \textbf{MoRo}, a generative framework for robust, efficient motion reconstruction from videos, which builds on recent advances in Masked Generative Transformers~\cite{devlin2019bert, motionbert2022, chang2022maskgit, zhang2023generating, guo2024momask, pinyoanuntapong2024mmm}. Namely, MoRo reformulates motion reconstruction as a video-conditioned generative task via masked modeling. 
Masked modeling with transformers has been widely adopted in text~\cite{devlin2019bert}, image~\cite{chang2022maskgit}, and motion~\cite{guo2024momask, pinyoanuntapong2024mmm, motionbert2022} generation. By randomly masking sequence segments, the model learns to reconstruct missing parts - an intuitive fit for handling occlusions. 
Unlike optimization- or diffusion-based methods~\cite{zhang2024rohm, phasemp, rempe2021humor}, which are slow and initialization-sensitive, masked modeling can enable efficient, end-to-end inference. 
While prior works such as GenHMR~\cite{saleem2024genhmr} and MEGA~\cite{fiche2025mega} apply this paradigm to single-frame mesh recovery, extending it to video is far more challenging: it requires not only resolving per-frame ambiguities but also modeling long-term dynamics across local and global pose spaces while remaining faithful to visual evidence under severe occlusions.


Directly learning the video-to-motion mapping under body occlusions as in~\cite{fiche2024vq} is challenging due to the scarcity of paired video–motion data. To address this, we decompose the learning process across diverse modalities spanning motion, image, and video datasets, and integrate them into a unified framework with end-to-end inference.
Following~\cite{fiche2024vq}, we represent 3D human meshes by discrete local pose tokens using a pre-trained Vector Quantized Variational Autoencoder (VQ-VAE)~\cite{oord2018neural}.
To recover motion from missing observations, it is crucial to model natural human dynamics. We begin by training a trajectory-aware motion prior on large-scale MoCap datasets~\cite{AMASS} with masked modeling, where the model jointly denoises a noisy input root trajectory and predicts missing local pose tokens. 
To overcome the limited pose diversity in MoCap, we then train an image-conditioned pose prior on large-scale image–pose datasets~\cite{human36m, mpi-inf-3dhp, mpii, coco} for pose reconstruction, while the image encoder of this prior also estimates a coarse global trajectory that serves as input to the motion prior. 
Finally, we fine-tune a video-conditioned masked motion transformer — combining the pretrained motion prior, pretrained image-conditioned pose prior, and a cross-modality decoder — on video datasets~\cite{zhang2022egobody, bedlam} via masked modeling, enabling the recovery of missing pose tokens and denoising of the global trajectory conditioned on video evidence.
A multi-step inference process iteratively recovers pose tokens from video evidence while refining the global trajectory.
Unlike prior occlusion-handling methods~\cite{phasemp, zhang2024rohm, rempe2021humor, zhang2021learning} that overlook visual context in motion prior learning, \mname{} unifies learning across diverse datasets and modalities in a single end-to-end framework, eliminating reliance on preprocessing and efficiently leveraging multi-modality priors to enhance robustness for motion recovery under occlusions.

In summary, our contributions are: 
1) \mname{}, a novel generative framework that leverages masked modeling for robust and efficient motion recovery from monocular videos; 
2) a cross-modality learning scheme that fuses multi-modal priors learnt across motion, image and video data, effectively learning a video-conditioned motion distribution.

Extensive evaluations show that \mname{} significantly outperforms state-of-the-art methods in both reconstruction accuracy and motion realism in challenging occlusion cases, while achieving comparable performance in non-occluded scenarios. 

\section{Related Work}
\label{sec:related}

\myparagraph{Human mesh recovery (HMR) from a single image} has seen significant progress in recent years. We can distinguish regression-based methods~\cite{goel2023humans,cho2022cross,Choi_2020_ECCV_Pose2Mesh,kanazawa2018end,Kocabas_PARE_2021,Kocabas_SPEC_2021,kolotouros2019cmr,li2021hybrik,li2022cliff,lin2021end-to-end,Moon_2020_ECCV_I2L-MeshNet,omran2018neural,xu2019denserac,zhang2021body,zhou2021monocular, kolotouros2021probabilistic, sarandi2024neural}, optimization-based methods~\cite{bogo2016smpl,fang2021reconstructing,lassner2017unite,SMPL-X:2019} and hybrid methods~\cite{kolotouros2019learning,song2020lgd}. 
Most methods regress SMPL~\cite{loper2015smpl} or SMPL-X~\cite{SMPL-X:2019} parameters, while others predict non-parametric mesh vertices~\cite{cho2022cross,Choi_2020_ECCV_Pose2Mesh,lin2021end-to-end,Moon_2020_ECCV_I2L-MeshNet} or arbitrary human volume points~\cite{sarandi2024neural} from images.
Recently, VQ-HPS~\cite{fiche2024vq} and TokenHMR~\cite{dwivedi_cvpr2024_tokenhmr} reformulate HMR from continuous regression to discrete classification by tokenizing human poses, showing improved accuracy.
HMR methods vary in focus: most aim for higher accuracy in generic scenarios, some enhance camera modeling~\cite{li2022cliff, Kocabas_SPEC_2021, patel2024camerahmr}, and others handle occlusions and truncations~\cite{kolotouros2021probabilistic, Kocabas_PARE_2021, wang2025prompthmr}. 
Building on tokenized pose representations, MEGA~\cite{fiche2025mega} and GenHMR~\cite{saleem2024genhmr} employ generative masked modeling to resolve pose ambiguities, producing multiple hypotheses from a single image.
However, these approaches remain limited to static images and cannot model temporal correlations.

\myparagraph{Human motion reconstruction from videos} aims at estimating plausible 3D human motion from frames. 

Early regression-based methods~\cite{you2023co, luo20203d, cheng19occlusion, choi2020beyond, kanazawa2019learning, nam2023cyclic, wei2022capturing} primarily predict local motion in the camera space without modeling the global trajectory, thus exhibiting motion artifacts.
Other optimization-based methods~\cite{rempe2021humor,phasemp,zhang2021learning} refine noisy per-frame estimates using motion priors and/or scene constraints, improving robustness under occlusions. However, they are slow, sensitive to local minima, and require extensive manual tuning. Moreover, their reliance on noisy per-frame estimates makes them fragile when the initialization is unreliable.
More recently, diffusion-based approaches such as RoHM~\cite{zhang2024rohm} tackle motion reconstruction under occlusions by conditioning on partial observations, yet they still rely on per-frame initialization and remain too slow for real-time use. 
In contrast, we propose to leverage the generative masked modeling framework to enable end-to-end and real-time inference.
Another recent line of work addresses dynamic camera scenarios. Some train regressors~\cite{wham:cvpr:2024, shen2024world}, while others integrate motion priors with SLAM-based reconstructions in optimization frameworks~\cite{ye2023slahmr,kocabas2024pace, li2024coin} to jointly estimate human and camera trajectories. However, when applied to videos with occlusions even under static cameras, these methods struggle to robustly reconstruct consistent motion (as shown in Sec.~\ref{sec:results}).

\myparagraph{Generative masked modeling.}
Masked modeling, initially introduced in BERT~\cite{devlin2019bert} for language tasks, was later adapted to vision through masked autoencoders~\cite{MaskedAutoencoders2021}, where models learn to reconstruct masked tokens from visible context. Building on this idea, masked generative modeling extends the paradigm by starting from a fully masked sequence and progressively generating tokens in fixed steps~\cite{chang2022maskgit,Chang2023MuseTG}. It has been applied to human motion generation~\cite{pinyoanuntapong2024controlmm, guo2024momask, pinyoanuntapong2024mmm, jiang2024motiongpt}, achieving state-of-the-art performance while being significantly faster than diffusion-based methods.

Recent works like GenHMR~\cite{saleem2024genhmr} and MEGA~\cite{fiche2025mega} extend the idea to human mesh recovery to generate multiple pose hypotheses from a single image, demonstrating its effectiveness when dealing with amgituities. Still, these methods are limited to static images. 
In contrast, we further extend the generative masked modeling framework to the video domain, reconstructing natural human motions from videos under occlusions.

\section{Motion Representation}
\label{sec:motion-repr}

\myparagraph{SMPL-X~\cite{SMPL-X:2019}} is a parametric body model that represents the 3D human body as a function $\mathcal{M}(\boldsymbol{\gamma}, \boldsymbol{\Phi}, \boldsymbol{\theta}, \boldsymbol{\beta}, \boldsymbol{{\theta}}_h, \boldsymbol{\phi})$, which returns a triangle mesh $\mathcal{M}$ with 10,475 vertices. It is parameterized by global translation $\boldsymbol{\gamma}$, global orientation $\boldsymbol{\Phi}$, body pose $\boldsymbol{\theta}$, body shape $\boldsymbol{\beta}$, hand pose $\boldsymbol{{\theta}_h}$ and facial expression $\boldsymbol{\phi}$. In this paper we consider only the main body parts while omitting $\boldsymbol{{\theta}}_h$ and $\boldsymbol{\phi}$.

\myparagraph{3D Body Mesh Tokenization.}
Following prior works~\cite{dwivedi_cvpr2024_tokenhmr,fiche2024vq, fiche2025mega}, we utilize a tokenized representation of the human
mesh. 
A local pose tokenizer is pre-trained to learn a discrete latent representation for the human mesh, adopting the convolutional autoencoder architecture from Mesh-VQ-VAE~\cite{fiche2024vq}. 
Given a SMPL-X mesh $\boldsymbol{v} \in \mathbb{R}^{10475\times 3}$ in local coordinates (setting the global orientation and translation to zero to disentangle global trajectory and local pose), the pose tokenizer encoder maps it into latent embeddings $\boldsymbol{z} \in \mathbb{R}^{P\times L}$, where $P=87$ is the number of tokens and $L=9$ is the dimension of each token. 
Each latent embedding $\boldsymbol{z}_i$ is then quantized into a discrete token $\tilde{\boldsymbol{z}}_i$ by finding its nearest neighbor in the codebook $\mathcal{C}$ of size 512, as $\tilde{\boldsymbol{z}}_i = \argmin_{\boldsymbol{c}_k \in \mathcal{C}}\|\tilde{\boldsymbol{z}_i}-\boldsymbol{c}_k\|_2^2$. The quantized tokens $\tilde{\boldsymbol{z}}$ are mapped back to a human mesh $\tilde{\boldsymbol{V}}$ by a symmetric decoder.
The local pose tokenizer is trained on AMASS~\cite{AMASS}, BEDLAM~\cite{bedlam} and MOYO~\cite{MOYO}, providing a strong prior on plausible human meshes.

\myparagraph{Motion representation.}
We represent a motion sequence of $T$ frames by $\boldsymbol{X} = (\boldsymbol{R}, \boldsymbol{Z})$, where $\boldsymbol{R} \in \mathbb{R}^{T\times 9}$ and $\boldsymbol{Z} \in \mathbb{R}^{T\times P\times L}$ denote the pelvis global trajectory, and quantized local body tokens, respectively. 
For frame $t$, the global trajectory $\boldsymbol{R}^t$ consists of the SMPL-X global orientation $\boldsymbol{\Phi} \in \mathbb{R}^6$ in 6D representation~\cite{zhou2019continuity} and the translation $\boldsymbol{\gamma} \in \mathbb{R}^3$.
The tokenized local body pose $\boldsymbol{Z}^t \in \mathbb{R}^{P\times L}$ is obtained from the pose tokenizer, consisting of $P$ discrete pose tokens with the dimension of each token as $L$.

\section{Method}
\label{sec:methods}

\begin{figure*}[t]
    \includegraphics[width=1.\textwidth]{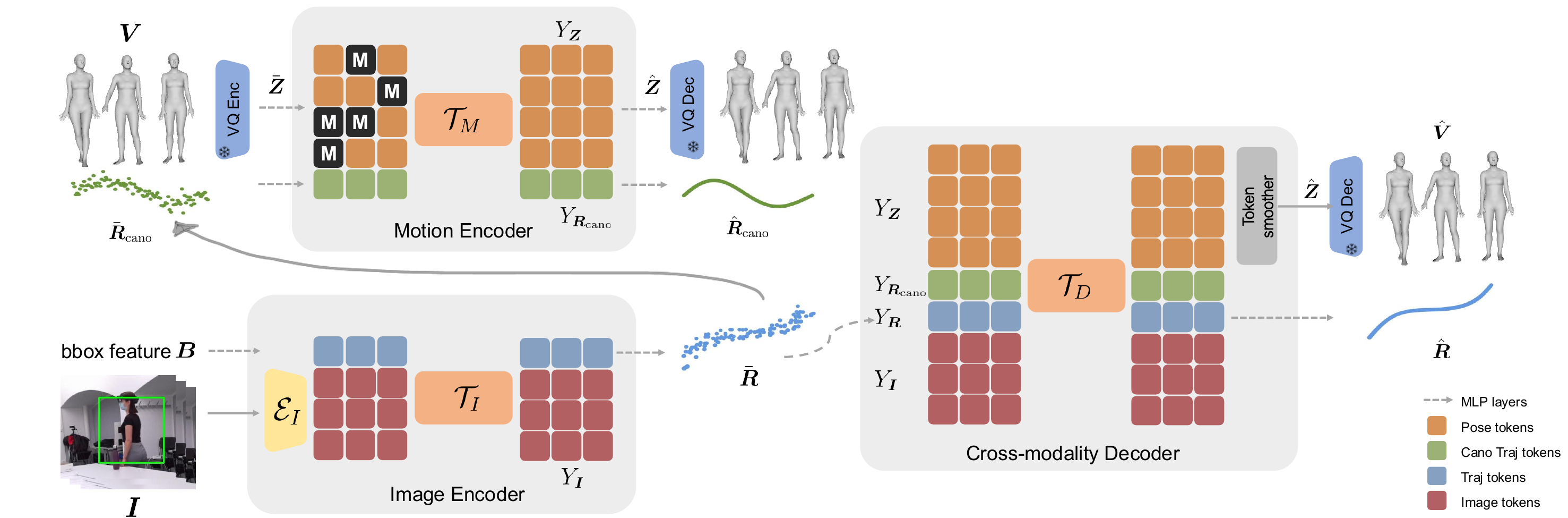}
    \caption{\textbf{Overview of our masked transformer, }which consists of three main components: the image encoder, the motion encoder and the decoder.
    Given a monocular video sequence, we utilize the image encoder to extract per-frame image features and estimate a coarse global trajectory, which is canonicalized and serves as the input to the motion encoder (Sec.~\ref{sec:image-encoder}). Along with masked local pose tokens, the motion encoder encodes a trajectory-aware motion prior via recovering the complete local pose tokens and denoising the global trajectory (Sec.~\ref{sec:motion-encoder}). The cross-modality decoder fuses the intermediate feature from both encoders via a spatial-temporal transformer to refine the camera-space global trajectory and predict a conditional categorical distribution for sampling the local pose tokens, which are then smoothed for enhanced motion realism (Sec.~\ref{sec:full-model}). 
    }
    \label{fig:overview}
\end{figure*}

We introduce \mname{}, a novel generative masked modeling framework for 3D human motion recovery from monocular videos under body occlusions. 
Given a monocular video $\boldsymbol{I}$ with $F$ frames captured by a static camera,
\mname{} aims to learn a conditional distribution of the human motion $p(\boldsymbol{X}| \boldsymbol{I})$.

\mname{} features three main components: an image encoder pretrained on large-scale image-pose datasets for visual conditioning and image-conditioned pose prior learning (Sec.~\ref{sec:image-encoder}), a motion encoder pretrained on large-scale MoCap datasets via masked modeling for motion prior learning (Sec.~\ref{sec:motion-encoder}), and a cross-modal decoder finetuned on video-motion data to fuse cross-modality information and efficiently recovering the human motion from videos (Sec.~\ref{sec:full-model}). 
A multi-step inference schedule iteratively predicts pose tokens and refines the global trajectory (Sec.~\ref{sec:inference}), further improving motion realism under occlusions.
The multi-stage training scheme is explained in Sec.~\ref{sec:training}.
Fig.~\ref{fig:overview} shows an overview of the proposed approach.

\subsection{Per-frame Image Conditioning} 
\label{sec:image-encoder}
The image encoder processes each frame to extract image features and estimate a coarse global trajectory.
A ViT-H/16~\cite{dosovitskiy2020image} backbone $\mathcal{E}_I$ initialized with pretrained weights from ViTPose~\cite{xu2022vitpose} encodes the cropped video frames $\boldsymbol{I}$ into image tokens $\mathcal{E}_I(\boldsymbol{I}) \in \mathbb{R}^{F\times 192 \times 1280}$, where 192 denotes the number of image tokens.
Inspired by~\cite{li2022cliff, zhang2023probabilistic}, the additional bounding box feature $\boldsymbol{B} = (\boldsymbol{b}_x,\boldsymbol{b}_y,\boldsymbol{s}) / \boldsymbol{f} \in \mathbb{R}^{F\times 3}$ is utilized to better infer the global positional information, 
where $\boldsymbol{b}_x,\boldsymbol{b}_y$ denote the bounding box center x-y coordinates relative to the principle point, $\boldsymbol{s}$ is the bounding box size in the original full image, and $\boldsymbol{f}$ is the focal length.

Tokens from $\mathcal{E}_I(\boldsymbol{I})$ and $\boldsymbol{B}$ are projected to the same latent dimension of 1024 by linear layers, concatenated and fed into a transformer network $\mathcal{T}_I$ to further model the visual context:
\begin{equation}
\begin{aligned}
    Y_{\boldsymbol{I}}, \bar{\boldsymbol{R}} &= \mathcal{T}_I(\mathcal{E}_I(\boldsymbol{I}),\boldsymbol{B}), \\ 
\end{aligned}
\label{eq:image-encoder}
\end{equation}

where $Y_{\boldsymbol{I}} \in \mathbb{R}^{F\times 192 \times 1024}$ denotes the encoded latent image feature, serving as the visual conditioning for the cross-modal decoder.

The weak-perspective camera parameters $\bar{\boldsymbol{R}}_{\textrm{crop}}$ in the cropped view are obtained by applying a linear layer to the latent feature at the bounding box token position. 
These parameters are then converted back to the full camera space using the CLIFF~\cite{li2022cliff} transformation and temporally stacked to form the coarse global trajectory initialization $\bar{\boldsymbol{R}}$.

Note that in the image encoder, each frame is processed separately without incorporating temporal information.

Specifically, the image encoder is combined with a pose decoder (architecturally identical to the cross-modal decoder in Sec.~\ref{sec:full-model} without modeling temporal information) and pre-trained on image–pose datasets~\cite{human36m, coco, mpi-inf-3dhp, mpii} to learn an image-conditioned pose prior via body pose reconstruction. This prior captures diverse poses present in image datasets but absent in video datasets, thereby facilitating video-conditioned motion learning in later stages.

\subsection{Trajectory-aware Motion Prior}
\label{sec:motion-encoder}
Directly recovering global human motion from image features leads to degraded motion quality under occlusions (see Sec.~\ref{sec:ablation}). To address this, we introduce a data-driven motion prior learned from the AMASS motion capture dataset~\cite{AMASS}, which models natural human dynamics from partial observations, improving temporal consistency and robustness to occlusions while reducing reliance on large-scale paired video–motion data.
Provided the noisy global trajectory $\bar{\boldsymbol{R}}$ predicted by the image encoder, inspired by the insight that human local pose is strongly correlated with its movement in the global space, we design the motion prior in a trajectory-aware manner to further model the strong inter-dependencies between local pose and global trajectory.

We transform $\bar{\boldsymbol{R}}$ into a canonicalized coordinate system in which each frame is represented by its motion toward the next frame, yielding the canonicalized global trajectory $\bar{\boldsymbol{R}}_\textrm{cano}$. 
For each frame $t$, the canonicalized global orientation $\bar{\boldsymbol{\Phi}}_\textrm{cano}^t$ and translation $\bar{\boldsymbol{\gamma}}_\textrm{cano}^t$ are computed as:

\begin{equation}
\begin{aligned}
    \bar{\boldsymbol{\Phi}}_\textrm{cano}^t &=
    \left(\bar{\boldsymbol{\Phi}}^{t}\right)^{-1} \bar{\boldsymbol{\Phi}}^{t+1}, \\
    \bar{\boldsymbol{\gamma}}_\textrm{cano}^t &
    = \left(\bar{\boldsymbol{\Phi}}^{t}\right)^{-1}(\bar{\boldsymbol{\gamma}}^{t+1} - \bar{\boldsymbol{\gamma}}^{t}),
\end{aligned}
\label{eq:canonicalization}
\end{equation}
This canonicalization makes the global trajectory independent of the coordinate system and sequence length, which is crucial for motion modeling.
We further apply a binary mask $\boldsymbol{M}\in\{0,1\}^{F\times P}$ to partially mask local pose tokens $\boldsymbol{Z}$ following the paradigm of masked modeling, resulting in the corrupted motion $(\bar{\boldsymbol{R}}_\textrm{cano}, \bar{\boldsymbol{Z}})$ where $\bar{\boldsymbol{Z}}=\boldsymbol{M} \odot \boldsymbol{Z}$. For motion pretraining on AMASS, $\bar{\boldsymbol{R}}_\textrm{cano}$ is obtained by manually corrupting the clean global trajectory by adding Gaussian noise to the body orientation and translation (see Supp.~Mat. for details).

A motion transformer $\mathcal{T}_M$ aims to recover the complete pose tokens and denoise the global trajectory simultaneously:
\begin{equation}
\begin{aligned}
    \hat{\boldsymbol{Z}}, \hat{\boldsymbol{R}}_\textrm{cano}, Y_{\boldsymbol{Z}}, Y_{\boldsymbol{R}_\textrm{cano}} &= \mathcal{T}_M(\bar{\boldsymbol{Z}},\bar{\boldsymbol{R}}_\textrm{cano}), \\
\end{aligned}
\label{eq:motion-encoder}
\end{equation}
where $\hat{\boldsymbol{Z}}$ and $\hat{\boldsymbol{R}}_\textrm{cano}$ denote the reconstructed pose tokens and trajectory, respectively. $Y_{\boldsymbol{Z}}$ and $Y_{\boldsymbol{R}_\textrm{cano}}$ are their corresponding latent features encoded by $\mathcal{T}_M$, and later fed to the cross-modal decoder (Sec.~\ref{sec:full-model}) for video-conditioned motion recovery. 
During motion pretraining, the recovered local pose tokens $\hat{\boldsymbol{Z}}$ are further mapped back to the original SMPL-X mesh sequence by the pose tokenizer for self-supervision in the vertex space.

Unlike previous works~\cite{zhang2024rohm, rempe2021humor, phasemp} where motion priors solely model motion itself, our proposed prior also acts as the motion encoder in the video-conditioned masked transformer and can be fine-tuned with video data, providing strong knowledge of natural human dynamics while conditioning on video inputs.

\subsection{Video-conditioned Masked Transformer}
\label{sec:full-model}

Given the per-frame image features, the pre-trained motion prior and image-conditioned pose prior, the video-conditioned masked transformer further incorporates a spatial–temporal transformer $\mathcal{T}_\mathcal{D}$ as the cross-modal decoder to fuse multi-modality information from both image and motion encoders to recover global motion. Following the same masked modeling strategy for training the motion prior, it predicts local pose tokens and the clean global body trajectory conditioning on visual observations.

Firstly, the image encoder predicts per-frame image features $Y_{\boldsymbol{I}}$ and the coarse global trajectory $\bar{\boldsymbol{R}}$ (Eq.~\ref{eq:image-encoder}). 
The canonicalized trajectory $\bar{\boldsymbol{R}}_\textrm{cano}$ obtained from $\bar{\boldsymbol{R}}$, together with partially masked pose tokens, are then processed by the motion encoder to produce the pose features $Y_{\boldsymbol{Z}}$ and trajectory features $Y_{\boldsymbol{R}_{\textrm{cano}}}$ (Eq.~\ref{eq:motion-encoder}). 
Finally, transformer $\mathcal{T}_{D}$ models the spatial-temporal correlations among the multi-modal features from image, pose and trajectory, and generate the complete pose token sequence $\hat{\boldsymbol{Z}}$ and a refined global trajectory $\hat{\boldsymbol{R}}$:
\begin{equation}
\begin{aligned}
    \hat{\boldsymbol{Z}}, \hat{\boldsymbol{R}} &= \mathcal{T}_\mathcal{D}
    ( Y_{\boldsymbol{Z}}, Y_{\boldsymbol{R}_{\textrm{cano}}}, Y_{I}, Y_{\boldsymbol{R}}) , \\
\end{aligned}
\end{equation}
where $Y_{\boldsymbol{R}}$ is obtained by encoding the estimated global trajectory $\bar{\boldsymbol{R}}$ from Eq.~\ref{eq:image-encoder} by a linear layer. 

The predicted pose tokens $\hat{\boldsymbol{Z}}$ are decoded by the pose tokenizer into the original SMPL-X mesh space, and then combined with the reconstructed global trajectory $\hat{\boldsymbol{R}}$ to produce the final reconstructed motion in the global space. 

\myparagraph{Pose token smoother network.} 
Due to the discretization during tokenization, the generated motion derived from per-frame pose tokens still exhibits some jittering artifacts. To address this, we inject a learnable smoother network $\mathcal{F}_{\text{smoother}}$ to map the discrete latent representation $\hat{\boldsymbol{Z}}$ picked from the codebook into a continuous representation, before decoding it into the canonical mesh. $\mathcal{F}_{\text{smoother}}$ is implemented as a 2-layer MLP, efficiently alleviating the jittering artifacts (Sec.~\ref{sec:ablation}).

\myparagraph{Architectures.}
The image encoder $\mathcal{T}_I$ employs a transformer encoder structure following~\cite{vaswani2017attention}.
Both the motion encoder $\mathcal{T}_M$ and the cross-modal decoder $\mathcal{T}_D$ adopt the spatial-temporal transformer architecture DSTFormer from~\cite{motionbert2022}. 
In addition, we leverage Rotary Positional Embedding (RoPE)~\cite{su2024roformer} by computing the temporal attention based on relative temporal positions, enabling \mname{} to handle sequences with variable length during inference.

\subsection{Inference}
\label{sec:inference}
At inference time, the model iteratively recovers masked tokens based on uncertainty of each predicted pose token.

The image encoder is first executed once to extract image features and predict a coarse global trajectory $\bar{\boldsymbol{R}}$ by Eq.~\ref{eq:image-encoder}. 
In the first inference iteration, fully masked pose tokens together with canonicalized $\bar{\boldsymbol{R}}_{\textrm{cano}}$ are fed into the motion encoder $\mathcal{T}_M$ and decoder $\mathcal{T}_D$ to generate the complete pose tokens and a refined global trajectory $\hat{\boldsymbol{R}}$. 
We then retain the top-$K$ pose tokens with the highest confidence, along with the refined global trajectory, as input for the next iteration, while the remaining tokens are re-masked for regeneration.

For each pose token, the confidence refers to the predicted logits after the softmax layer. 
The refined global trajectory $\hat{\boldsymbol{R}}$ at each iteration will be canonicalized and fed to the motion encoder to be refined in the next iteration. 
The process repeats until all tokens are recovered. We adopt $T=5$ as the number of inference iteration steps. The final pose tokens are decoded into the SMPL-X mesh and transformed to global coordinates using predicted trajectory from the last iteration.

\subsection{Multi-stage Training}
\label{sec:training}
In order to strike the balance between faithfully recovering motion from available image evidence and generating realistic motions for occluded body parts, the proposed model is trained in a progressive manner, spanning datasets with different annotation modalities.

\myparagraph{Motion Pretraining.} 
The motion encoder is pretrained on AMASS~\cite{AMASS}. 
In addition to random masking adopted by previous works~\cite{fiche2025mega, saleem2024genhmr}, we introduce spatial and temporal masking strategies that either mask all tokens in selected frames or mask specific tokens across all frames during training, better reflecting real-world scenarios where occlusions are typically continuous in space and time.

\myparagraph{Image Pretraining.} 
We pretrain the image encoder and the cross-modal decoder on standard image datasets - including Human3.6M~\cite{human36m}, MPI-INF-3DHP~\cite{mpi-inf-3dhp}, COCO~\cite{coco} and MPII~\cite{mpii} - to improve generalization to diverse body poses, which are less represented in video datasets. During image pretraining, motion-related features ($Y_{\boldsymbol{Z}}, Y_{\boldsymbol{R}_\textrm{cano}}$) are simply masked out in the decoder.

\myparagraph{Video Fine-tuning.} 
Finally, the full model is fine-tuned on two monocular video-motion datasets, EgoBody~\cite{zhang2022egobody} and BEDLAM~\cite{bedlam}. The spatial-temporal cross-modal decoder leverages the pre-trained motion prior information and image-pose prior information to further model the correlations among multiple modalities from visual inputs, global trajectory and local motion.

\myparagraph{Confidence-guided masking.}
Alongside random masking, we adopt a confidence-guided masking strategy during video fine-tuning. Starting with fully masked inputs, we perform one inference step, re-mask a subset of tokens according to their predicted confidence as in the iterative inference (Sec.~\ref{sec:inference}), and run a subsequent prediction round.
This reduces the train–test gap and enables the model to recover tokens from imperfect inputs in multi-step inference.

\myparagraph{Training objective.}
\mname{} is trained with the cross entropy loss $\mathcal{L}_\textrm{ce}$ for the local pose token classification, local 3D mesh vertex loss $\mathcal{L}_{V_{3D}}$, global trajectory loss $\mathcal{L}_\textrm{traj}$, global 3D joint position loss $\mathcal{L}_{J_{3D}}$ and velocity loss $\mathcal{L}_{\dot{J}_{3D}}$, 2D keypoint reprojection loss $\mathcal{L}_{J_{2D}}$ and foot skating loss $\mathcal{L}_\textrm{fs}$:
\begin{equation}
    \mathcal{L} = \mathcal{L}_{\text{ce}} 
    + \mathcal{L}_{V_{3D}}
    +\mathcal{L}_{\text{traj}}
    +\mathcal{L}_{J_{3D}} 
    + \mathcal{L}_{\dot{J}_{3D}} 
    + \mathcal{L}_{J_{2D}}
    +\mathcal{L}_{\text{fs}},
\end{equation}

where the global joint losses $\mathcal{L}_{J_{3D}},\mathcal{L}_{\dot{J}_{3D}}$ are computed from multiple global trajectory predictions $\bar{\boldsymbol{R}}, \hat{\boldsymbol{R}}_\textrm{cano}, \hat{\boldsymbol{R}}$ from our model 
and $\mathcal{L}_{\text{fs}}$ penalizes the foot velocity if it exceeds a certain threshold and is in contact with ground to reduce foot skating artifacts. 
Each loss term is weighted by corresponding weights and applied only when it is applicable in the pretraining stages. Please see Supp.~Mat. for more details.

\section{Experiments}
\label{sec:experiments}

\subsection{Datasets}
\mname{} is trained on datasets across multiple modalities as described in Sec.~\ref{sec:training}, and evaluated on \textbf{EgoBody}~\cite{zhang2022egobody} and \textbf{RICH}~\cite{Huang:CVPR:2022}. 
Both EgoBody and RICH capture human motions interacting with various 3D environments, recording multi-modal data streams including third-person view RGB videos, with the human motion annotated with SMPL-X parameters. 
EgoBody includes a notable amount of occlusion scenarios, and we evaluate the proposed method on a subset of EgoBody third-view videos with severe body occlusions following~\cite{zhang2024rohm}, excluding sequences in the EgoBody training split. This curated subset is denoted as \textit{EgoBody-Occ} and consists of 17 video sequences with a total of 23055 frames. 
RICH, on the other hand, has relatively uncluttered scenes, resulting in few occlusions in videos. It has been a standard evaluation dataset for evaluating video-based motion reconstruction~\cite{shen2024world, kocabas2024pace, wang2024tram, wham:cvpr:2024}. We evaluate on 191 test sequences to assess model performance in non-occluded scenarios.

\subsection{Implementation Details}
The predicted SMPL-X mesh vertices from our method are fitted to SMPL-X parameters using a fast fitting algorithm~\cite{sarandi24nlf} for evaluation.
%
We use the same bounding box detections across all methods for fair comparison. On EgoBody, bounding boxes are derived from OpenPose 2D keypoints~\cite{openpose}, excluding keypoints with confidence below 0.2. On RICH, we adopt the preprocessed bounding boxes provided by~\cite{shen2024world}. 
Ground truth camera intrinsics is employed in all methods.
%
We perform extreme cropping augmentation~\cite{dwivedi_cvpr2024_tokenhmr} by randomly cropping human body parts from images to further improve the model's robustness to truncated bounding boxes.

\subsection{Evaluation Metrics}

\myparagraph{Accuracy.}
The local pose and shape accuracy is evaluated via Mean Per Joint Position Error (\textit{MPJPE} in $mm$), Procrustes-aligned MPJPE (\textit{PA-MPJPE} in $mm$), and Per Vertex Error (\textit{PVE} in $mm$).
Following~\cite{zhang2024rohm}, we report MPJPE for full-body (\textit{-all}), visible joints (\textit{-vis}) and occluded (\textit{-occ}) joints separately on EgoBody-occ. 
%
For global-space reconstruction, prior works~\cite{wham:cvpr:2024, wang2025prompthmr, shen2024world, wang2024tram, yuan2022glamr} often report World MPJPE, which aligns predicted and ground-truth motions by the first frame over each 100-frame segments, thereby underestimating global translation errors in long sequences. Instead, we report Global MPJPE (\textit{GMPJPE} in $mm$) and Root Translation Error (\textit{RTE} in \%)~\cite{shen2024world} to evaluate long-term global accuracy.

\myparagraph{Motion Quality.}
We report metrics on motion smoothness and foot sliding of the reconstructed motion to evaluate the motion plausibility. Consistent with prior works~\cite{wham:cvpr:2024, wang2025prompthmr, shen2024world, wang2024tram, yuan2022glamr}, we report the local acceleration error (Accel, in $m/s^2$), motion jitter (in $m/s^3$), and foot sliding (in $mm$). 
Additionally, we find that global acceleration error (G-Accel, in $m/s^2$) better reflects the motion smoothness globally.

\subsection{Baselines}
\label{sec:baseline}
We compare \mname{} against (1) per-frame pose estimation methods: MEGA~\cite{fiche2025mega}, TokenHMR~\cite{dwivedi_cvpr2024_tokenhmr}, PromptHMR~\cite{wang2025prompthmr}\footnotemark[1], and (2) video-based motion reconstruction methods: RoHM~\cite{zhang2024rohm}, WHAM~\cite{wham:cvpr:2024}, GVHMR~\cite{shen2024world}. 
RoHM is a diffusion-based method that relies on per-frame 3D body pose initializations and precomputed occlusion masks. 
WHAM and GVHMR, though designed for dynamic-camera settings, also applies to static cameras by fixing camera extrinsics to identity. Both output a camera-space trajectory and a world-grounded trajectory, which should differ only by a rigid transform under the static camera setup; however, we find them inconsistent in practice due to an additional refinement step on the world-grounded trajectory -
the world-grounded trajectory achieves better motion realism but degrades video–motion alignment (Fig.~\ref{fig:qualitative}, row 3), while the camera-space trajectory aligns better with video but yields lower motion quality. 
They evaluate per-frame metrics (PA-MPJPE, MPJPE, PVE) using camera-space predictions and global metrics (RTE, Jitter, Foot-Sliding) using world-grounded predictions, whereas a single model should ideally produce unified outputs consistent across both camera and global space.
We therefore report results separately for each prediction, denoted by \textit{-Cam} and \textit{-World}.
Please refer to Supp.~Mat. for more details. 

\footnotetext[1]
{While PromptHMR proposes a video-based variant (PromptHMR-Vid), the code for reproducing its result on RICH is unavailable. Thus we only evaluate the per-frame model. We will provide evaluation for PromptHMR-Vid in future versions once the evaluation code is provided.}

\subsection{Results}
\label{sec:results}
\myparagraph{Performance on videos with occlusions.}
Tab.~\ref{tab:egobody-occ} shows the quantitative results on EgoBody-Occ.
Our method consistently surpasses baselines in both accuracy and motion quality, demonstrating strong robustness under occlusions.

For local joint accuracy, our model outperforms all baselines on both visible and occluded joints, achieving a \textbf{16/31\%} MPJPE improvement over the best baseline PromptHMR for visible/occluded body parts, respectively. Although PromptHMR is relatively robust to various bounding box sizes by encoding full-image context and using the bounding box only as a spatial prompt, our method still surpasses it.
In terms of global reconstruction, our model exceeds the best baseline RoHM by a large margin, achieving \textbf{58\%} better global joint reconstruction (GMPJPE). RoHM suffers from poor local pose accuracy (with a high PA-MPJPE) since it ignores visual input during the motion prior learning, whereas our method effectively addresses visual evidence and the motion prior within one unified framework.

\begin{figure}[h]
  \centering
  \includegraphics[width=0.48\textwidth]{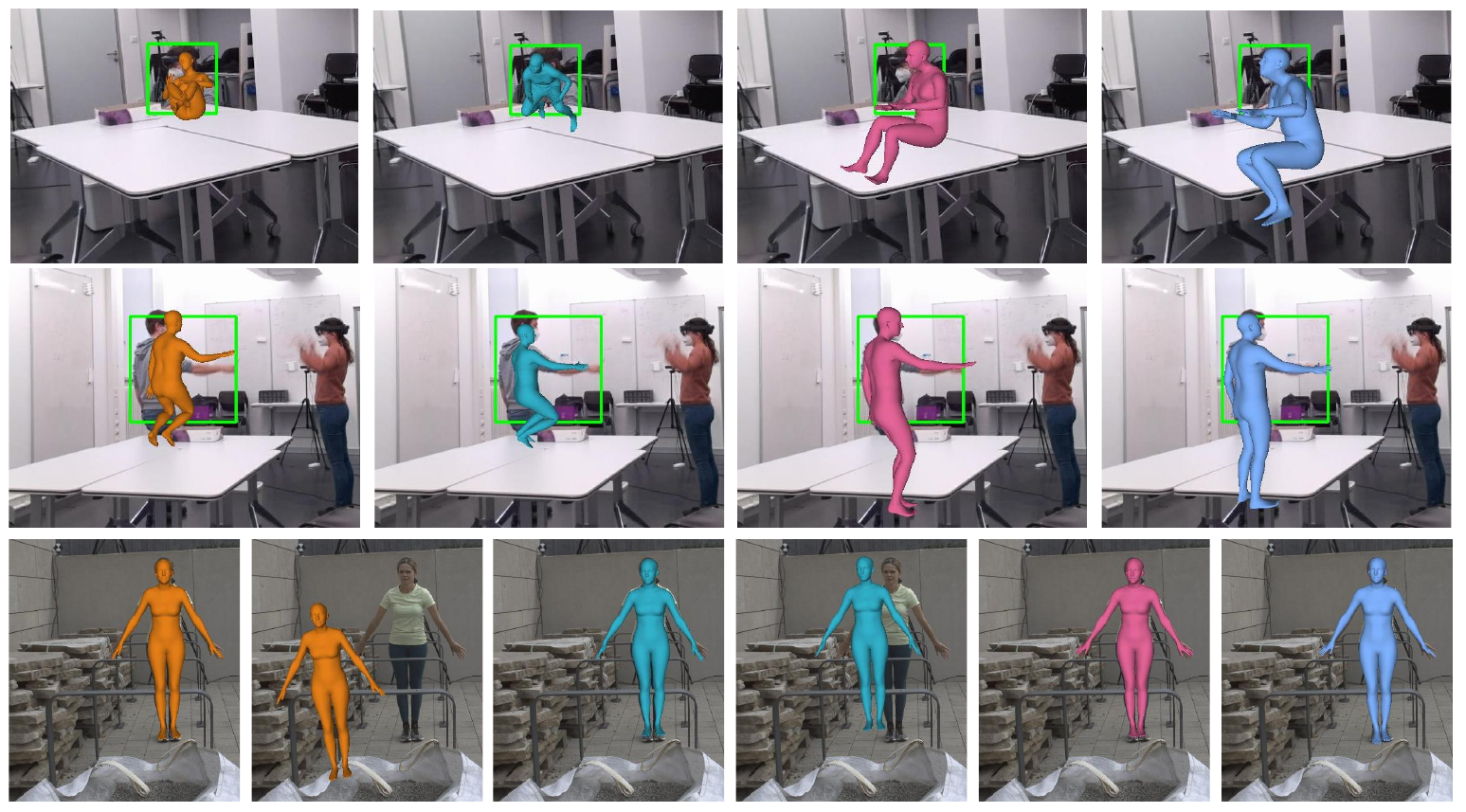}
  \caption{\small \textbf{Qualitative examples on EgoBody (row 1, 2) and RICH (row 3)}. For row 1-2, from left to right corresponds to \textcolor{orange}{WHAM-Cam}, \textcolor{blue-green}{GVHMR-Cam}, \textcolor{pink}{PromptHMR} and \textcolor{blue}{ours}. For row 3, from left to right corresponds to \textcolor{orange}{WHAM-Cam}, \textcolor{orange}{WHAM-World}, \textcolor{blue-green}{GVHMR-Cam}, \textcolor{blue-green}{GVHMR-World}, \textcolor{pink}{PromptHMR} and \textcolor{blue}{ours}.}
  \label{fig:qualitative}
\end{figure}

Regarding motion realism, our method produces remarkably plausible motion with the least jitters and the second least foot sliding. RoHM generates fixed-length clips, leading to temporal discontinuities for long sequences, while our RoPE-based~\cite{su2024roformer} transformer maintains consistency over arbitrary-length videos by attending to local 60-frame contexts. 
As mentioned in Sec.~\ref{sec:baseline}, WHAM and GVHMR produce two sets of inconsistent outputs: camera-space predictions (-Cam) yield better per-frame accuracy but suffer from motion jitter, while world-grounded predictions (-World) improve motion realism by additional neural network blocks but drift from from visual evidence in the image plane (see Fig.~\ref{fig:qualitative}). That is to say, WHAM and GVHMR struggle to simultaneously deliver accurate per-frame pose and smooth, physically plausible motion with a unified output.
In contrast, our method leverage the motion prior to enforce temporal consistency and the video-conditioned decoder to enforce the predicted motion to align with the visual cues, and both integrated to an end-to-end framework, producing one unified trajectory in the global space, achieving a good balance between motion realism and image alignment.

Qualitative results in Fig.~\ref{fig:qualitative} (row 1, 2) further demonstrate that our method yields substantially improved robustness under occlusions compared with the baselines.

\myparagraph{Performance on occlusion-free videos.}
We further quantitatively evaluate \mname{} in an occlusion-free scenario on RICH (Tab.~\ref{tab:rich}).
Our method delivers comparable results as baselines while yielding more plausible motion (lower Accel, G-Accel, and Jitter).
Fig.~\ref{fig:qualitative} (row 2) illustrates that our reconstructions align closely with the video input.

\begin{table*}[tb]
\centering
\small
\scalebox{0.92}{
\begin{tabular}{@{}llccccc|cc|cccc@{}}
\toprule[1pt]
& \multirow{2}{*}{\textbf{Method}} & \multirow{2}{*}{\textbf{PA-MPJPE}$\downarrow$} & \multicolumn{3}{c}{\textbf{MPJPE}$\downarrow$}  & \multirow{2}{*}{\textbf{PVE}$\downarrow$}  & \multirow{2}{*}{\textbf{GMPJPE}$\downarrow$}  & \multirow{2}{*}{\textbf{RTE}$\downarrow$}  & \multirow{2}{*}{\textbf{Accel}$\downarrow$} & \multirow{2}{*}{\textbf{G-accel}$\downarrow$}  & \multirow{2}{*}{\textbf{Jitter}$\downarrow$} & \multirow{2}{*}{\textbf{Sliding}$\downarrow$} \\
& & & -\textit{all} & -\textit{vis} & -\textit{occ}  & & & & & & \\
\midrule
\multirow{3}{*}{\rotatebox[origin=c]{90}{\scriptsize per-frame}} & MEGA~\cite{fiche2025mega}   &  37.80  &  86.92  &  83.98  & 108.17  &  106.13  & - & - & - & - & - & - \\
& TokenHMR~\cite{dwivedi_cvpr2024_tokenhmr} & 38.07  & 76.38  & 72.94  & 101.30  & 94.71  & - & - & - & - & - & - \\
& PromptHMR~\cite{wang2025prompthmr} & \underline{35.00} & \underline{48.34}  & \underline{45.30}  & \underline{70.42}  & \underline{59.79}  & - & - & - & - & - & -\\
\midrule
\multirow{6}{*}{\rotatebox[origin=c]{90}{\scriptsize temporal-based}} & RoHM~\cite{zhang2024rohm} & 54.53  & 79.01  & 75.85  & 101.7  & 105.18  & \underline{308.8}  & \underline{2.23} & \underline{2.81}  & 3.78  & 12.74  & 3.28 \\
& WHAM~\cite{wham:cvpr:2024} -Cam & 44.20 & 82.03 & 77.33 & 116.1 & 98.46  & 745.23  & 5.18  & 3.27  & 115.68 & 626.52  & 72.78 \\
& WHAM~\cite{wham:cvpr:2024} -World  & 44.21 & 95.26 & 91.20 & 124.64 & 116.17 & 739.49  & 3.98  & 3.19  & \underline{3.36}   & \underline{10.27}   & \textbf{2.73} \\
& GVHMR~\cite{shen2024world} -Cam   & 48.85 & 71.00 & 64.95 & 114.87 & 83.60  & 877.26  & 3.53  & 3.53  & 81.76  & 441.84  & 52.49 \\
& GVHMR~\cite{shen2024world} -World   & 48.85 & 73.33 & 67.40 & 116.29 & 86.13  & 875.45  & 3.06  & 4.11  & 5.62   & 25.11   & 3.81 \\
& Ours  & \textbf{26.72} & \textbf{39.13} & \textbf{37.83} & \textbf{48.53}  & \textbf{50.25}  & \textbf{129.22}  & \textbf{0.58}  & \textbf{2.21}  & \textbf{2.15}   & \textbf{4.60}     & \underline{3.21} \\
\bottomrule
\end{tabular}
}
\caption{\textbf{Quantitative evaluation results on EgoBody-occ.}
The best / second best results are in \textbf{boldface}, and \underline{underlined}, respectively. For per-frame methods, we report only pelvis-aligned accuracy metrics since they lack global or temporal modeling by design.}
\label{tab:egobody-occ}
\end{table*} 

\begin{table*}[tb]
\centering
\small
\scalebox{0.92}{
\begin{tabular}{@{}llccc|cc|cccccc@{}}
\toprule[1pt]
& \textbf{Method} & \textbf{PA-MPJPE}$\downarrow$ & \textbf{MPJPE}$\downarrow$ & \textbf{PVE}$\downarrow$ & 
\textbf{GMPJPE}$\downarrow$ & \textbf{RTE}$\downarrow$ & \textbf{Accel}$\downarrow$ & \textbf{G-Accel}$\downarrow$ & \textbf{Jitter}$\downarrow$ & \textbf{Sliding}$\downarrow$ \\
\midrule
\multirow{3}{*}{\rotatebox[origin=c]{90}{\scriptsize per-frame}} & MEGA~\cite{fiche2025mega}      & 50.53 & 108.27 & 122.43 & - & - & - & - & - & - \\
& TokenHMR~\cite{dwivedi_cvpr2024_tokenhmr}    & 40.37 & 77.74  & 90.68  & -  & -  & - & - & - & - \\
& PromptHMR~\cite{wang2025prompthmr}     & \underline{38.17} & \textbf{58.56} & \textbf{67.25} & - & - & - & - & - & - \\
\midrule
\multirow{5}{*}{\rotatebox[origin=c]{90}{\scriptsize temporal-based}} & WHAM~\cite{wham:cvpr:2024} -Cam            & 44.53 & 79.80  & 90.65  & 557.06  & 2.94  & 5.67  & 49.41  & 258.94 & 33.29 \\
& WHAM~\cite{wham:cvpr:2024} -World          & 44.53 & 102.58 & 117.07 & 660.60  & 4.40  & 5.49  & 6.51   & 21.01  & \underline{3.97} \\
& GVHMR~\cite{shen2024world} -Cam           & 39.78 & \underline{66.07}  & \underline{75.71}  & \underline{520.51}  & 2.42  & \underline{4.15}  & 17.36 & 83.92  & 14.57 \\
& GVHMR~\cite{shen2024world} -World         & 39.78 & 73.72  & 83.85  & 553.66  & \underline{2.40}  & 4.40  & \underline{4.46}   & \underline{12.82}  & \textbf{2.99}  \\
& Ours & \textbf{35.37} & 74.00 & 84.47 & \textbf{378.43} & \textbf{1.45} & \textbf{3.71} & \textbf{3.59} & \textbf{5.90} & 4.86 \\
\bottomrule
\end{tabular}
}
\caption{\textbf{Quantitative evaluation results on RICH.}
The best / second best results are in \textbf{boldface}, and \underline{underlined}, respectively.
}
\label{tab:rich}
\end{table*}

\subsection{Ablation Studies}
\label{sec:ablation}
We conduct ablation studies on EgoBody-occ to examine the impact of key design choices and the number of inference steps.
As shown in Tab.~\ref{tab:ablation}, both motion realism and pose accuracy for occluded body parts improve notably when incorporating the motion prior (`ours' vs. `w/o ME'), highlighting its effectiveness. 
The motion encoder (Sec.~\ref{sec:motion-encoder}) not only enables masked modeling but, when pretrained on MoCap data, also better captures natural motion dynamics. 
%
The confidence-guided masking strategy (Sec.~\ref{sec:training}) further narrows the train–test gap, improving model's robustness during multi-step inference (`ours' vs. w/o `CGM').
Finally, the pose token smoother (Sec.~\ref{sec:full-model}) mitigates motion jitters resulted from discrete quantization in the tokenizer (`ours' vs. w/o $\mathcal{F}_{\text{smoother}}$').
Study on inference steps reveals that, increasing inference steps consistently reduces jitter and foot sliding, improving overall motion realism. Meanwhile , global pose accuracy peaks at $T=5$ steps, which we adopt as our final setup.


\begin{table}[tb]
\centering
\footnotesize
\scalebox{0.9}{
\begin{tabular}{@{}lcccccc@{}}
\toprule[1pt]
\multirow{2}{*}{\textbf{Method}} & 
\multicolumn{2}{c}{\textbf{MPJPE}$\downarrow$}  & 
\multirow{2}{*}{\textbf{GMPJPE}$\downarrow$}  & 
\multirow{2}{*}{\textbf{G-accel}$\downarrow$}  & 
\multirow{2}{*}{\textbf{Jitter}$\downarrow$} & 
\multirow{2}{*}{\textbf{Sliding}$\downarrow$} \\
 & -\textit{vis} & -\textit{occ}  & & & \\

\midrule
%
Ours   & \textbf{37.83} & \textbf{48.53} & \textbf{127.17} & \textbf{2.15} & \textbf{4.60} & \textbf{3.21} \\
w/o ME      & 39.77 & 55.89 & 134.31 & 3.66 & 14.20 & 5.92 \\
w/o CGM  & 40.32 & 53.90 & 136.38 & \underline{2.24} & \underline{5.13} & \textbf{3.21} \\
w/o $\mathcal{F}_{\text{smoother}}$   & \underline{38.66} & \underline{49.50} & 133.31 & 5.12 & 24.31 & 5.95 \\
\midrule
\multicolumn{7}{c}{Inference steps} \\
\midrule
T=1           & 38.27 & 51.10 & 129.22 & 2.32 & 6.12 & 3.57 \\
T=5           & \textbf{37.83} & \underline{48.53} & \textbf{127.17} & \textbf{2.15} & 4.60 & 3.21 \\ 
T=10          & \underline{37.85} & \textbf{48.51} & \underline{129.21} & \textbf{2.15} & \textbf{4.55} & \underline{3.14} \\
T=20          & 37.92 & 48.65 & 129.52 & \textbf{2.15} & \underline{4.58} & \textbf{3.10} \\

\bottomrule
\end{tabular}
}
\caption{\textbf{Ablation study on EgoBody-occ} for the motion encoder (`ME', Sec.~\ref{sec:motion-encoder}), pose token smoother ($\mathcal{F}_{\text{smoother}}$, Sec.~\ref{sec:full-model}), confidence-guided masking during training (`CGM', Sec.~\ref{sec:training}), and inference step numbers. 
The best / second best results are in \textbf{boldface}, and \underline{underlined}, respectively.}
\vspace{-0.2cm}
\label{tab:ablation}
\end{table}

\section{Conclusion}
\label{sec:conclusion}
We introduced \mname{}, a masked generative transformer framework for occlusion-robust human motion reconstruction from monocular video. 
MoRo leverages masked modeling and effectively consolidates multi-modal information across a set of heterogeneous datasets (MoCap, image-pose and video-motion data). By integrating a trajectory-aware motion prior and an image-conditioned pose prior into a video-conditioned generative transformer, MoRo recover temporally consistent human motion in global space in an end-to-end manner. Experiments show that MoRo outperforms state-of-the-art methods under occlusions while maintaining real-time performance, offering a practical solution for various downstream applications.

\myparagraph{Limitations and future work.} Despite its effectiveness, our method is currently restricted to static camera setups with known intrinsics, which limits its applicability to videos captured by moving cameras. In future work, we plan to incorporate techniques for modeling camera motion to extend \mname{} to dynamic camera scenarios. 

\myparagraph{Acknowledgements.} This work was supported as part of the Swiss AI initiative by a grant from the Swiss National Supercomputing Centre (CSCS) under project IDs \#36 on Alps, enabling large-scale training. We sincerely thank Muhammed Kocabas for his help with the PromptHMR codebase, and Korrawe Karunratanakul for insightful discussions.

\clearpage
{
    \small
    \bibliographystyle{ieeenat_fullname}
    \bibliography{main}
}


\end{document}